\documentclass[runningheads,a4paper]{llncs}
\usepackage{verbatim}
\usepackage{wrapfig}
\usepackage[table,xcdraw]{xcolor}
\usepackage{multirow}
\usepackage{amssymb}
\usepackage{graphicx}
\usepackage{rotating}
\usepackage{adjustbox}
\usepackage[utf8]{inputenc}
\usepackage{url}
\usepackage{amsmath}
\usepackage{enumitem}

\begin{document}

\mainmatter  

\title{Multi-Planar Deep Segmentation Networks for Cardiac Substructures from MRI and CT}

\titlerunning{Segmenting Cardiac Substructures with Deep Learning}

%
%

%


%
%
\author{Aliasghar Mortazi$^{1}$, Jeremy Burt$^{2}$, Ulas Bagci$^{1}$}
\authorrunning{Aliasghar Mortazi, Jeremy Burt, Ulas Bagci}
\institute{$^{1}$ Center for Research in Computer Vision (CRCV), University of Central Florida, Orlando, FL.\\$^{2}$ Diagnostic Radiology Department, Florida Hospital, Orlando, FL.\\ \emph{E-mail: a.mortazi@knights.ucf.edu}}

\toctitle{Lecture Notes in Computer Science}
\tocauthor{Authors' Instructions}
\maketitle

\begin{abstract}
Non-invasive detection of cardiovascular disorders from radiology scans requires quantitative image analysis of the heart and its substructures. There are well-established  measurements that radiologists use for diseases assessment such as ejection fraction, volume of four chambers, and myocardium mass.  These measurements are derived as outcomes of precise segmentation of the heart and its substructures. The aim of this paper is to  provide such measurements through an accurate image segmentation algorithm that automatically delineates seven  substructures of the heart from MRI and/or CT scans. Our proposed method is based on multi-planar deep convolutional neural networks (CNN) with an adaptive fusion strategy where we automatically  utilize complementary information from different planes of the 3D scans for improved delineations. For CT and MRI, we have separately designed three CNNs (the same architectural configuration) for three planes, and have trained the networks from scratch for voxel-wise labeling for the following cardiac structures: myocardium of left ventricle (Myo), left atrium (LA), left ventricle (LV), right atrium (RA), right ventricle (RV), ascending aorta (Ao), and main pulmonary artery (PA). We have evaluated the proposed method with 4-fold-cross-validation on the multi-modality whole heart segmentation challenge (MM-WHS 2017) dataset. The precision and dice index of 0.93 and 0.90, and 0.87 and 0.85 were achieved for CT and MR images, respectively. While a CT volume was segmented about 50 seconds, an MRI scan was segmented around 17 seconds with the GPUs/CUDA implementation.      

\end{abstract}

\section{Introduction}
According to the World Health Organization~\cite{who}, cardiovascular diseases (CVDs) are the first cause of death globally. About 17.7 million people died from CVDs in 2015, which was 31\% of total global deaths from diseases. Almost 7.4 million of these deaths were due to CVDs and about 6.7 million were due to the stroke. Extensive research and clinical applications have shown that both CT and MRI have vital roles in  non-invasive assessment of CVDs. CT is used more frequently than  MRI due to its fast acquisition and cheaper cost. On the other hand, MRI has an excellent soft tissue contrast and no ionizing radiation. However, most commercially available  image analysis methods have been either tuned for CT or MRI only. Furthermore, many studies are focused on only  one substructure of the heart (for instance, the left ventricle or left atrium).  Surprisingly, there is very little published research on  segmenting all substructures of the heart despite the fact that  clinically established markers rely on shape, volumetric, and tissue characterization of all the cardiac substructures.  Our study is concerned with this open problem from a machine learning perspective. We have investigated  architectural designs of  deep learning networks to solve multi-label  and multi-modality image segmentation challenges within the scope of a limited  GPU and imaging data.\\  

\noindent\textbf{Related Works.}
Literature related to cardiac image segmentation is vast. Among these works, atlas-based methods have been quite popular and favored for many years. For instance, multi-atlas based whole-heart segmentation using MRI and CT by~\cite{zhuang2016} and atlas propagation based method using prior information by~\cite{zhuang2008} are a few key examples. Despite their accuracy,  those methods  often lack efficiency  due to  heavy computations on the registration algorithms (e.g., from 13 minutes to 11 hours of computations reported in the literature).  Interested readers can find a survey paper on cardiac image segmentation methods in~\cite{review} for  a full list of methods and their comparative evaluations. 

More recently, deep learning based approaches are replacing the conventional methods in medical image segmentation fields in general, and  cardiac field in particular. For instance, in~\cite{mortazi2017}, a multi-planar deep learning has been utilized to segment LA and pulmonary veins from MR images. A recurrent fully convolutional neural network has been proposed to segment LV from MRI in~\cite{recurrent}. In a similar fashion, a deep learning algorithm combined with a deformable-model approach was used to segment LV from MRI~\cite{combined}. In~\cite{rvdeep}, RV segmentation has been accomplished through a joint localization and segmentation algorithm within a deep learning framework. To date, the majority of deep learning methods have segmented only one or two structures of the heart and constrained to only one modality, unlike what is presented herein.\\    

\noindent\textbf{Our Contributions.}
We have constructed a network structure similar to the one devised in~\cite{mortazi2017}, which segments the left atrium and proximal veins from MRI successfully.  In this paper, we have  extended this segmentation engine in several different ways as follows. (1) A deeper CNN has been utilized as compared to~\cite{mortazi2017}. (2) We have used both CT and MRI to test and evaluate the proposed system while Mortazi et al. used only MRI~\cite{mortazi2017}. (3) We have extended the binary segmentation problem into a multi-label segmentation problem. (4) We have devised a rank based adaptive fusion method to assess effective information from different planes for all delineated objects and select the best fusion strategies for highly accurate and efficient delineation results.

\section{Multi-Object Multi-Planar CNN (MO-MP-CNN)}
\begin{figure}[t]
\centering
\includegraphics[height=4cm,width=1\textwidth]{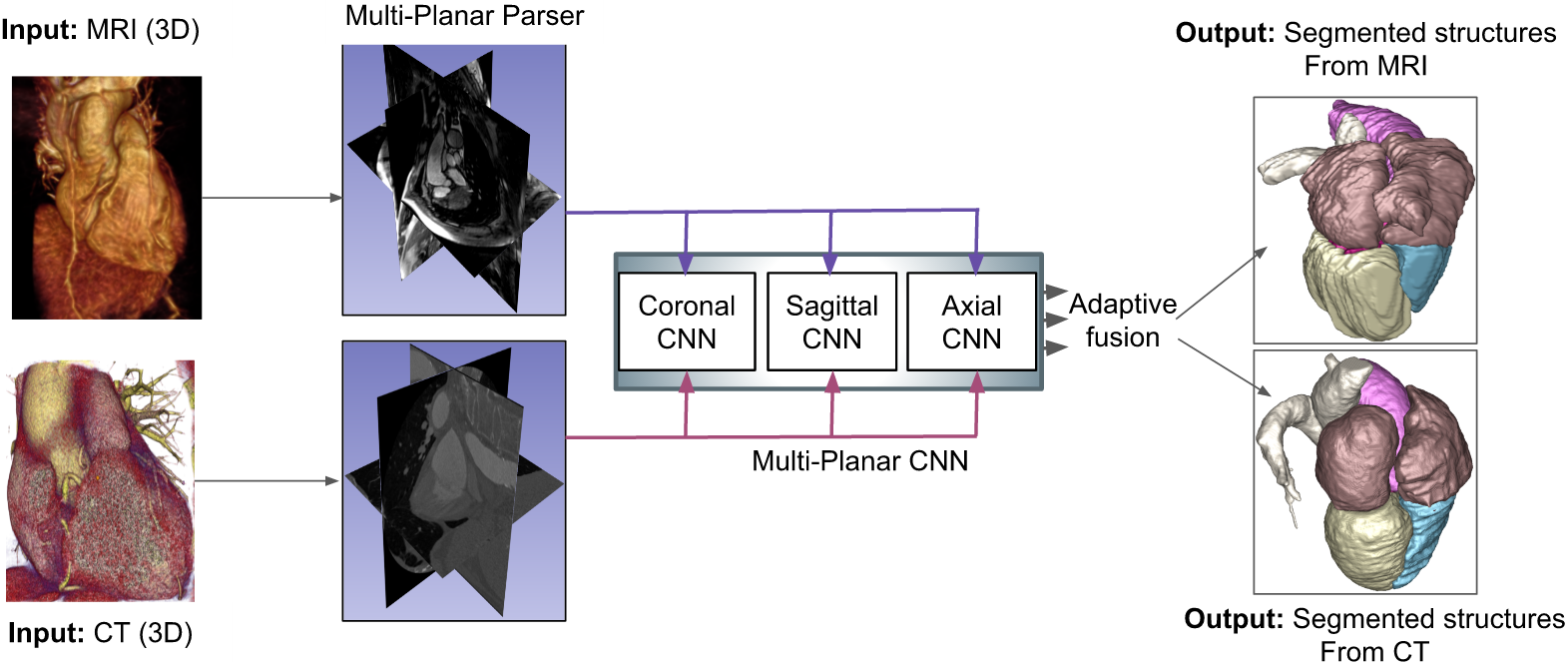}
\caption{Overall view of Mo-MP-CNN with adaptive fusion\label{fig:overall}}
\end{figure}

\vspace{-0.2 cm}
The proposed method is called multi-object multi-planar convolutional neural networks (MO-MP-CNN), and its modules are illustrated in Figure \ref{fig:overall}. MO-MP-CNN takes 3D CT or MR scans as an input and  parses it to three perpendicular planes: Axial(A), Coronal(C), and Sagittal(S). For each plane (and modality), a 2D CNN is  trained  to label  pixels.  CNNs have been trained from scratch to adapt into CT and MRI context. After training each of the 2D CNNs separately, adaptive fusion strategy is utilized by combining the probability maps of each of the CNNs. The details of the CNN and adaptive fusion method are explained in the following.\\   
 
\begin{figure}[h]
\centering
\begin{turn}{0}
\centering
\includegraphics[width=1\textwidth]{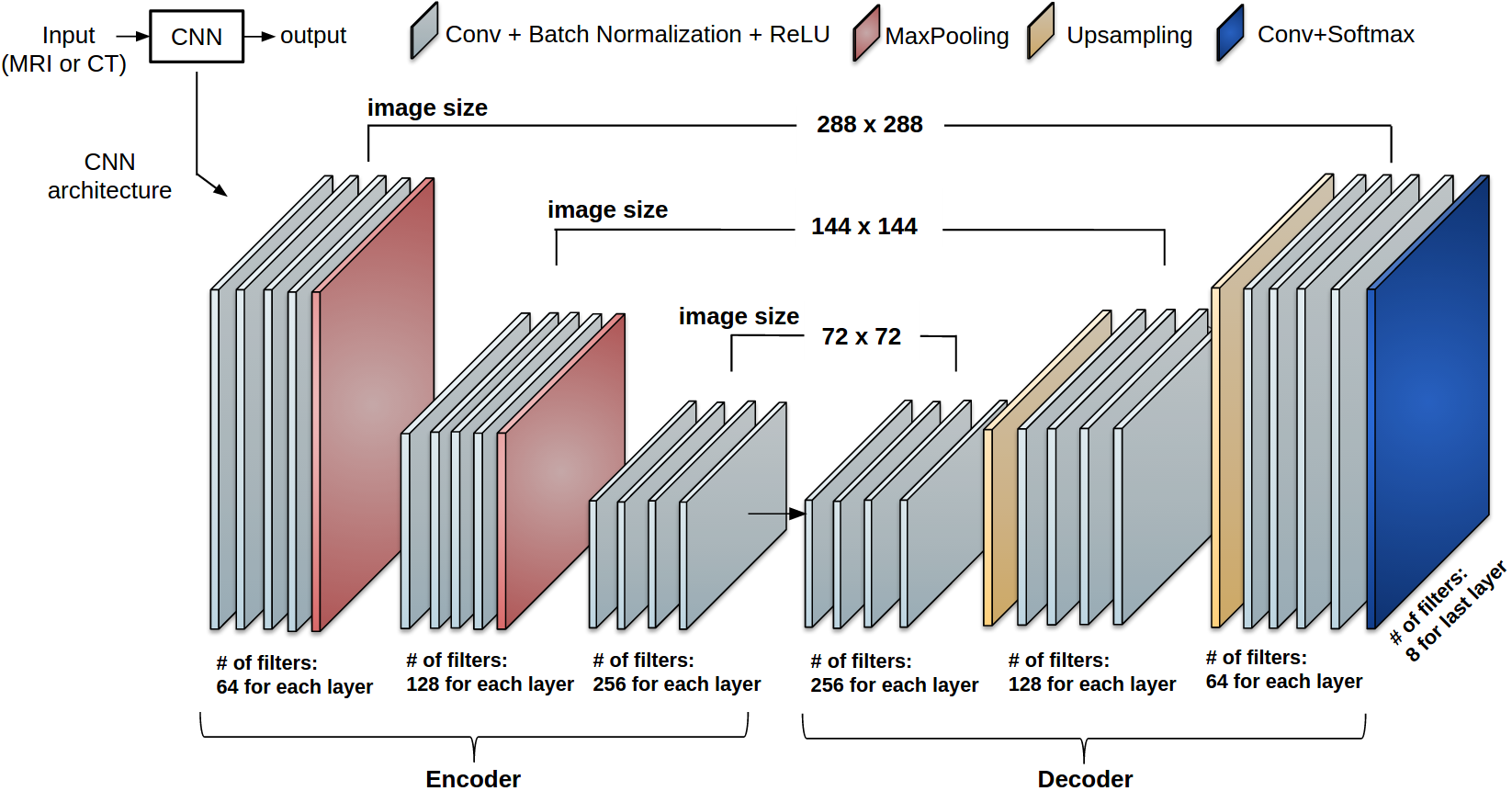}
\end{turn}
\caption{Details of the CNN architecture. Note that image size is not necessarily fixed for each plane's CNN.\label{fig:cnn}}
\vspace{-0.6 cm}
\end{figure}

\begin{wraptable}{r}{0.49\textwidth}
\vspace{-1 cm}
\caption{Data augmentation parameters.}
\begin{adjustbox}{max width=0.5\textwidth}

\begin{tabular}{|c|c|c|}
\hline
\multicolumn{3}{|c|}{\cellcolor[HTML]{C0C0C0}\textit{\textbf{Data augmentation}}}    \\ \hline
\textbf{Methods}         & \multicolumn{2}{c|}{\textbf{parameters}}                  \\ \hline
Zoom in                  & \multicolumn{2}{c|}{Scale $\varepsilon [1.1,1.3]$}                                  \\ \hline
Rotation                 & \multicolumn{2}{c|}{{$k\times45, k\, \varepsilon [-1,1]$ }}                                  \\ \hline
\multicolumn{3}{|c|}{\cellcolor[HTML]{C0C0C0}\textit{\textbf{Training images (CT)}}} \\ \hline
\textbf{CNN}             & \textbf{\# of images}        & \textbf{Image size}        \\ \hline
Sagittal                 & 40,960                       &350$\times$350                            \\
Axial                    & 21,417                       & 350$\times$350                            \\
Coronal                  &  40,960                      & 350$\times$350                           \\ \hline
\multicolumn{3}{|c|}{\cellcolor[HTML]{C0C0C0}\textit{\textbf{Training images (MRI)}}} \\ \hline
\textbf{CNN}             & \textbf{\# of images}        & \textbf{Image Size}        \\ \hline
Sagittal                 & 20,074                       &288$\times$288                            \\
Axial                    & 29,860                       &288$\times$160                            \\
Coronal                  & 19,404                       &288$\times$288                            \\ \hline

\end{tabular}
\end{adjustbox}
\vspace{-0.5 cm}
\end{wraptable}
\noindent\textbf{CNN network.} The proposed encoder-decoder based  network architecture is illustrated in Fig.\ref{fig:cnn}. Twelve \textit{convolution} layers have been used in encoder and decoder separately. In the encoder part, two \textit{max-pooling} layers have been used to reduce the dimension of the image by half and in decoder part two \textit{upsampling} layers (bilinear interpolation) have been used to get the image back to its original size. The size of all filters were set as $3\times3$. Each convolution layer is followed by a \textit{batch normalization} and \textit{Rectified Linear Unit (ReLU)} as an activation function. The number of filters in the last convolution layer is equal to the number of classes (i.e., 8 (background+7 objects)) and is followed by a \textit{ softmax} function to make a final probability map for each object. Similar to~\cite{mortazi2017}, the simplified z-loss~\cite{zloss} function has been used to train the network.  To provide a sufficient number of training images for the networks, data augmentation has been applied to the training images by rotation and zoom-in operations. The details of the augmentation and the number of data for each CNN are summarized in Table 1.

\subsubsection*{Multi-object adaptive fusion.} 

\begin{figure}[h]
\vspace{-0.4cm}
\centering
\includegraphics[width=1\textwidth]{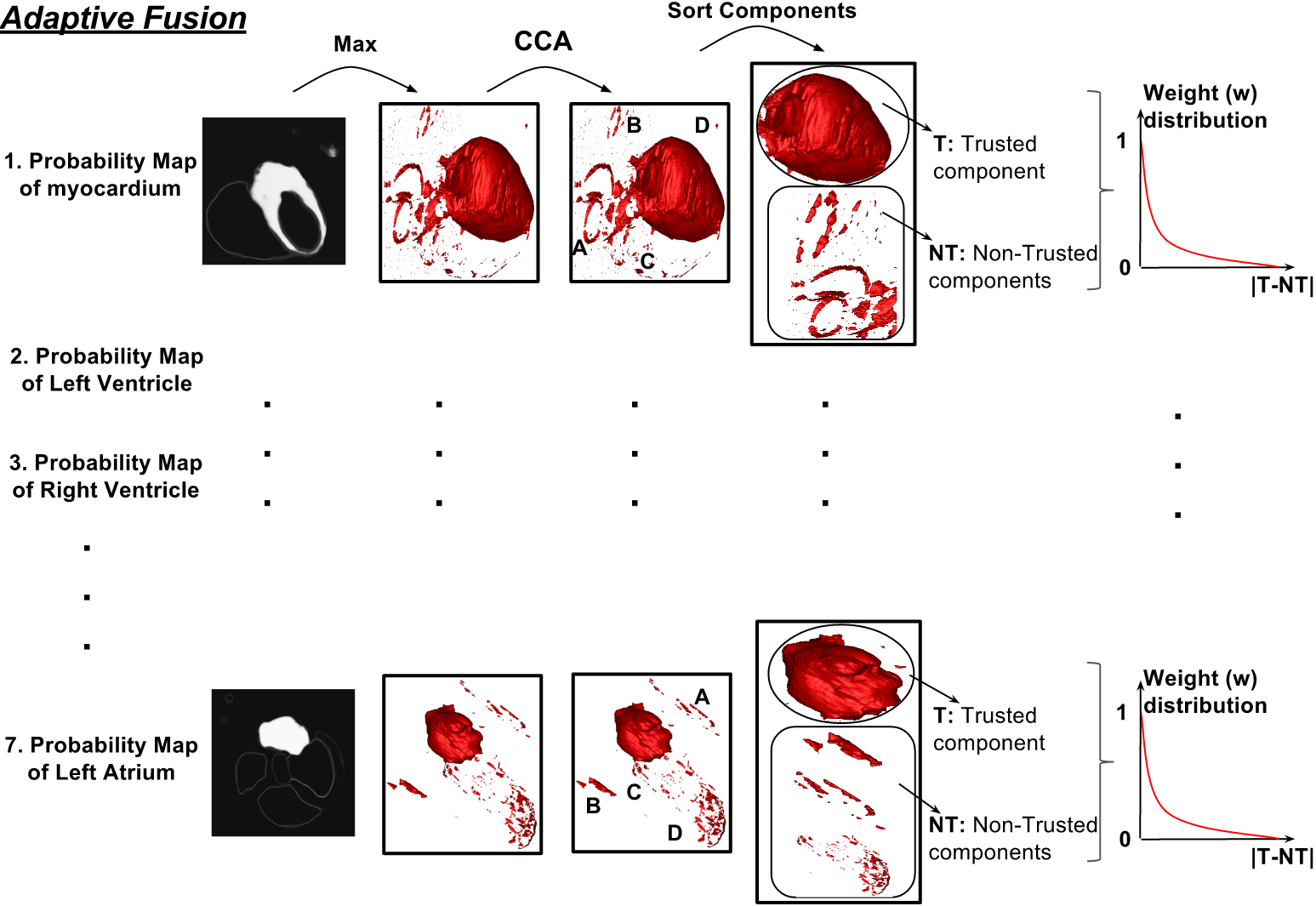}
\caption{Connected components obtained from each plane were computed and the residual volume (T-NT) was used to determine the strength for fusion with the other planes.\label{fig:fusion}}
\end{figure}

An adaptive fusion strategy has been extended in the way that it can be applied to multi-object segmentation instead of binary segmentation. Let $\mathbf{I}$ and $\mathbf{P}$ denote an input and output image pair, where output is the probability map of the CNN. Also, let the final segmentation be denoted as $\mathbf{o}$. As shown in Fig.\ref{fig:fusion}, $\mathbf{o}$ is obtained from the probability map $\mathbf{P}$ by taking the maximum probability of each pixel in all classes (labels). Then, a connected component analysis (CCA) is applied to $\mathbf{o}$ to select reliable and unreliable regions, where unreliable regions are considered to come from false positive findings. Although this approach gives a ``rough" estimation of the object, this information can well be used for assessing the quality of segmentations from different planes. If it is assumed that $\mathbf{n}$ is the number of classes (structures) in the images and $\mathbf{m}$ is the number of components in each class, then connected component analysis can be performed as follows: $CCA (\mathbf{o})=\{o_{11},\ldots,o_{nm} | \cup o_{ij} = \mathbf{o}, \text{ and } o_{11},\ldots,o_{nm} | \cap o_{ij}=\phi\}$. For each class $\mathbf{n}$, we can now assign reliability parameters (weights) to increase the influence of planes that have more reliable (trusted) segmentations as follows: $ w= \sum_ {i} \{max_{j} \{|o_{j}| \}\}/  \sum_{ij} |o_{ij}| $, where $w$ indicates a weight parameter. In our interpretation of the CCA, the difference between trusted and non-trusted regions have been used to guide the reliability of the segmentation process: the higher the difference is, the more reliable the segmentation is (See Fig.\ref{fig:fusion},  weight distribution w.r.t the difference). In test phase, we have simply used  those predetermined weights from the training stage.

\vspace{-0.8 cm}

\section{Experimental Results}
\noindent\textbf{Dataset and preprocessing:} For the experiments and evaluations of the proposed method, we used the STACOM 2017 for whole heart segmentation challenge dataset, containing 20 MR and 20 CT images for training (with ground-truth) and 40 test images without ground-truth for each modality. We performed a 4 fold cross-validation on the dataset such that 15 subjects were used for training and 5 subjects have been chosen for validation for each fold. The CT images were obtained from routine cardiac CT angiography and to cover the whole heart, extending from the upper abdomen to the aortic arch. Axial in-plane resolution was $0.78\times0.78$ mm and slice thickness was 1.6 mm. The MR images were acquired by using 3D balanced steady state free precession (b-SSFP) sequences, with about 2 mm acquisition resolution in each direction. In preprocessing step, anisotropic smoothing filtering was applied to both CT and MR images prior to segmentation. In addition, histogram matching was used for MR images to alleviate intensity non-standardness issues. 

\begin{figure}[h]
\vspace{-0.4cm}
\includegraphics[width=0.48\textwidth]{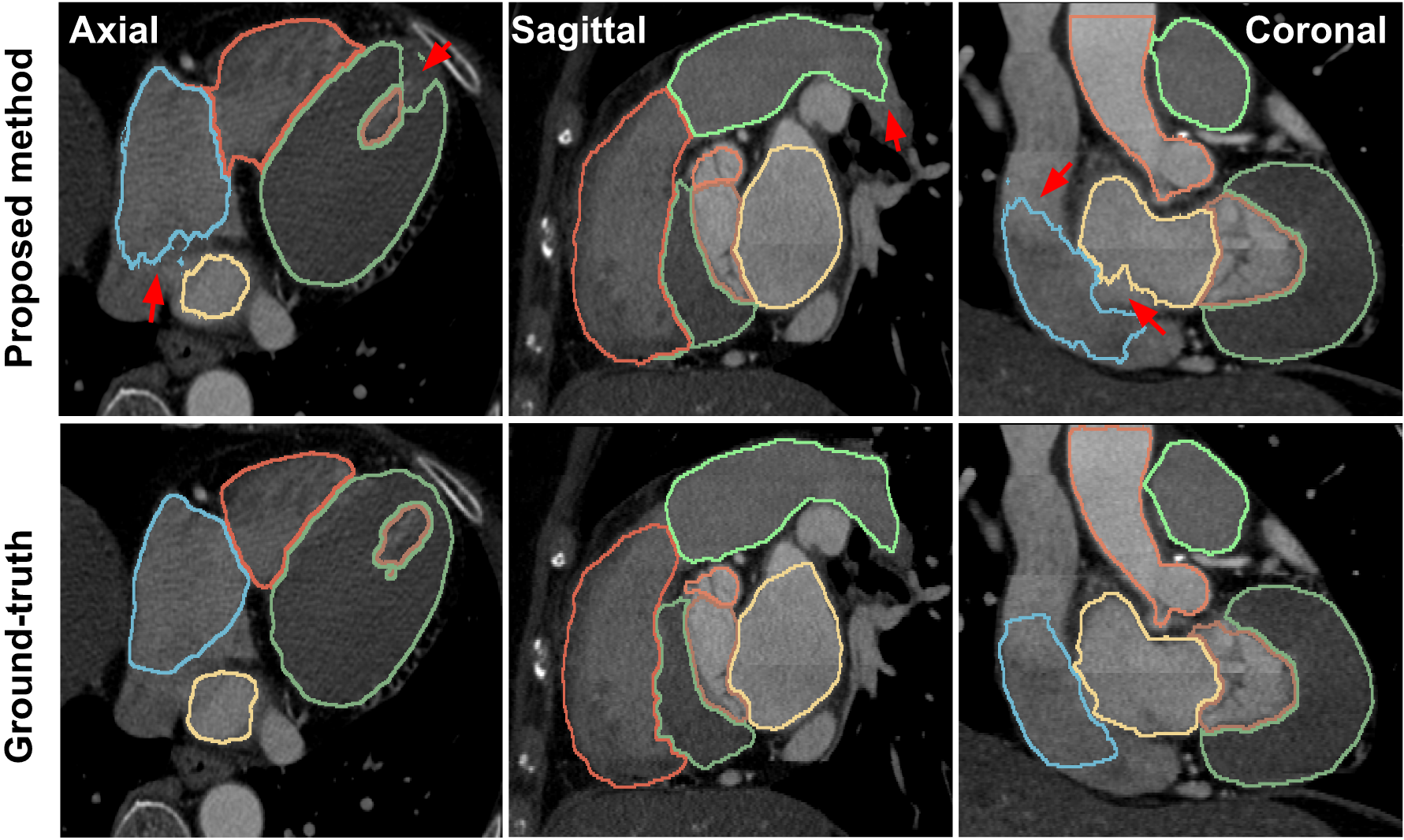}
\includegraphics[height= 3.5cm,width=0.51\textwidth]{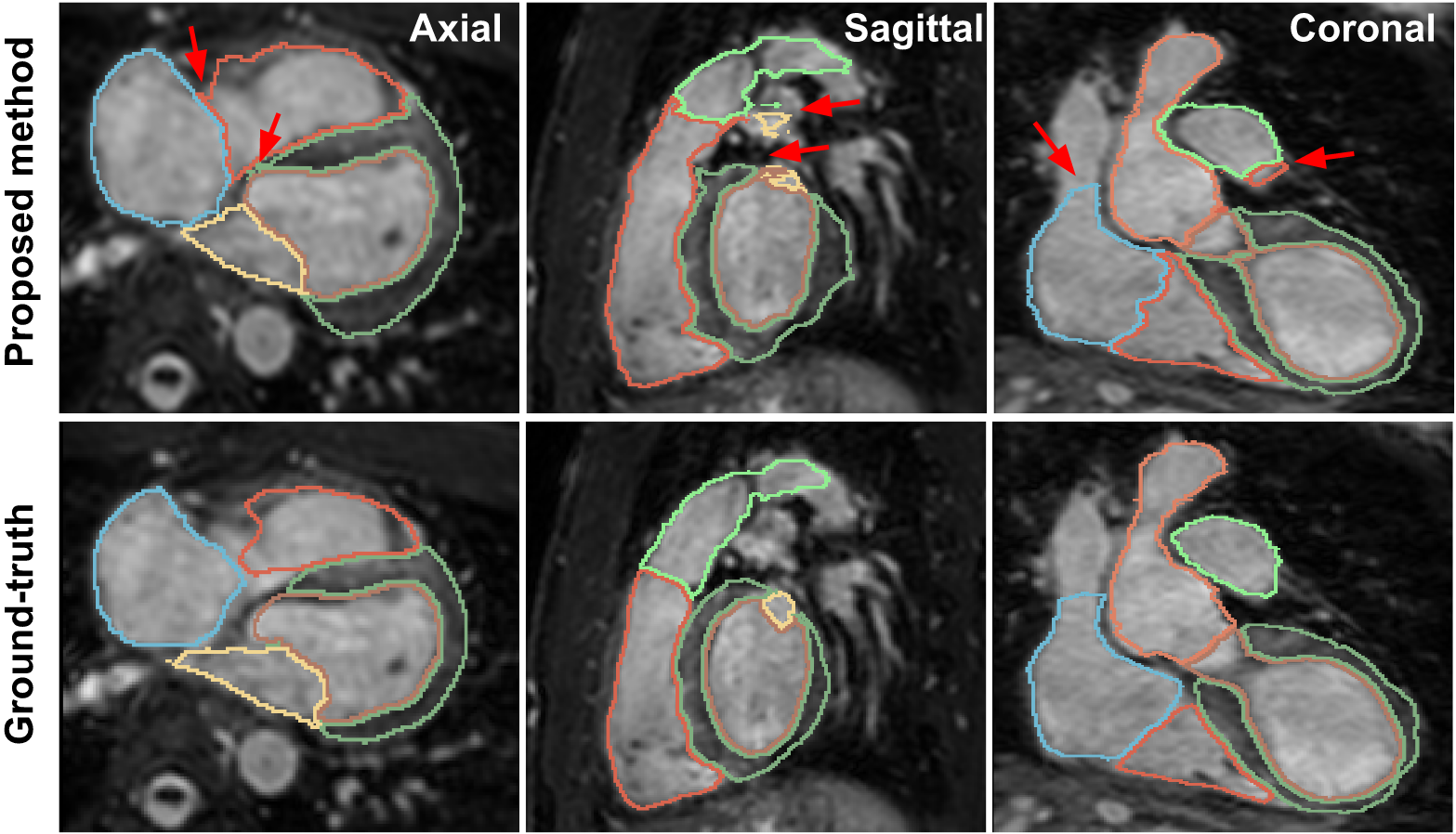} 
\includegraphics[width=0.48\textwidth]{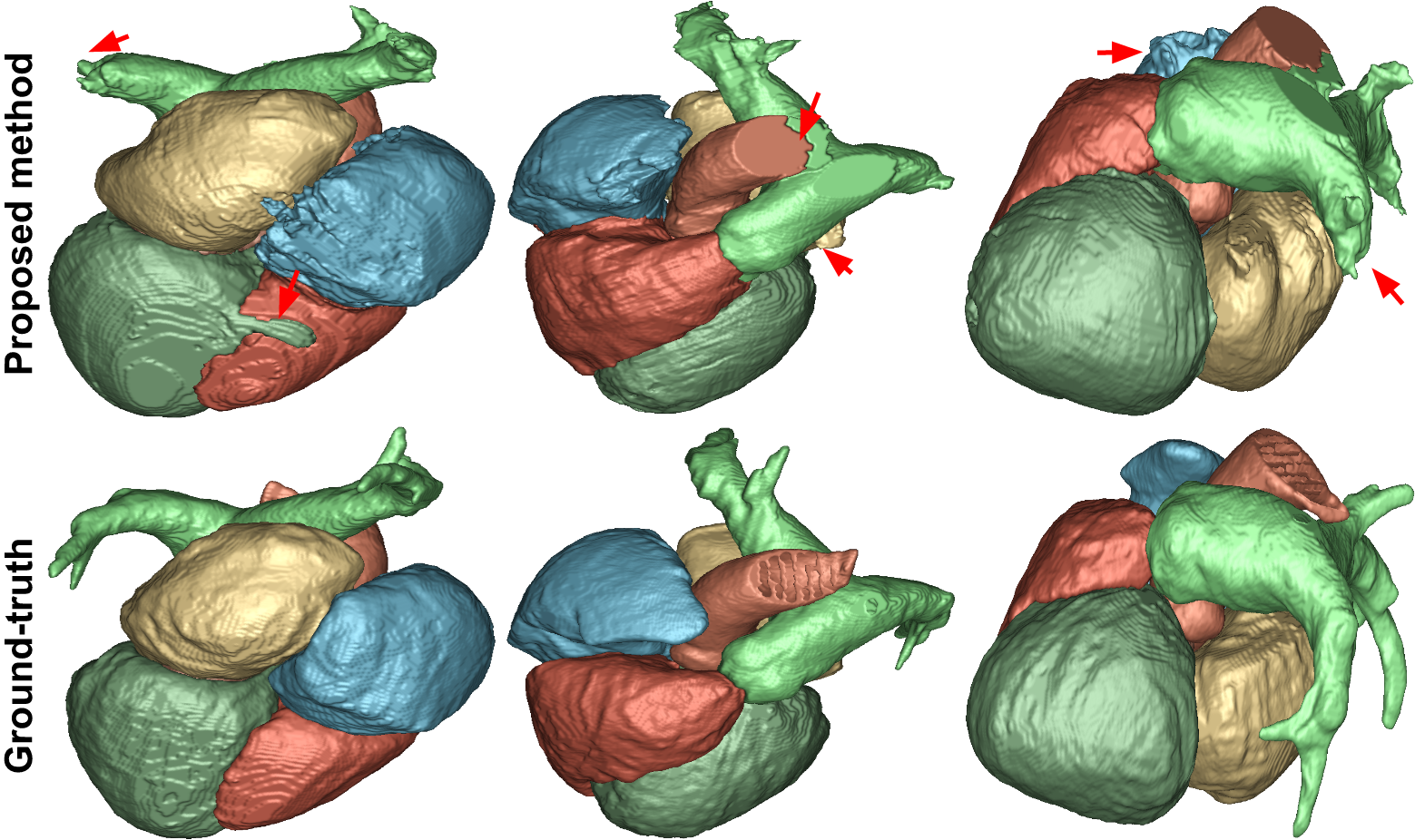}
\includegraphics[width=0.51\textwidth]{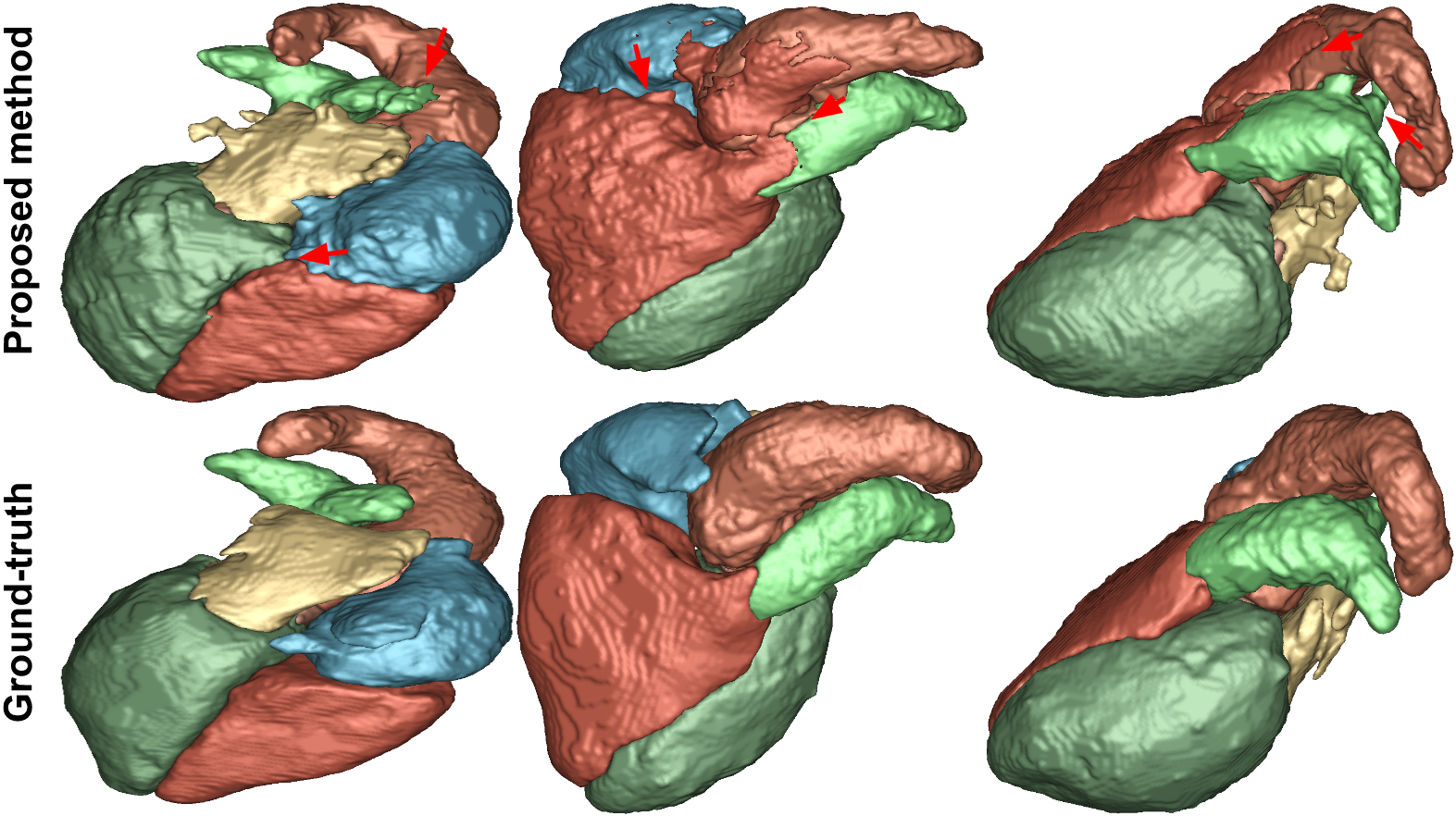}
\includegraphics[width=1\textwidth]{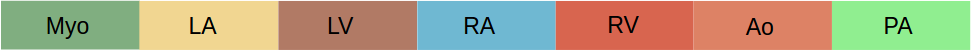}
\caption{First two rows show axial, sagittal, and coronal planes of the CT (first three columns) and MR images (last three columns), annotated cardiac structures, and their corresponding surface renditions (last two rows). Red arrows indicate some of mis-segmentations.\label{fig:qual} }
\vspace{-0.8cm}
\end{figure}


\begin{table}[!h]
\centering
\caption{Quantitative evaluations of the proposed segmentation method for both CT and MRI are summarized.}
\begin{adjustbox}{max height=14 cm,max width=1\textwidth}
\begin{tabular}{c|cccccccc|cccccccc|}
\cline{2-17}
                                           & \multicolumn{8}{c|}{{\cellcolor[HTML]{C0C0C0}\textit{\textbf{MRI}}}}                                                                                                                                                                                     & \multicolumn{8}{c|}{{\cellcolor[HTML]{C0C0C0}\textit{\textbf{CT}}} }                                                                                                                                                                                      \\ \hline
\multicolumn{1}{|c|}{\textbf{Structures:}}       & \textit{Myo}             & \textit{LA}              & \textit{LV}              & \textit{RA}             & \textit{RV}              & \textit{Aorta}           & \textit{PA}              & \textit{WHS}              & \textit{Myo}            & \textit{LA}              & \textit{LV}              & \textit{RA}              & \textit{RV}              & \textit{Aorta}           & \textit{PA}              & \textit{WHS}              \\ \hline
\multicolumn{1}{|c|}{\textbf{Sensitivity}} & 0.816                    & 0.856                    & 0.928                    & 0.827                   & 0.878                    & 0.728                    & 0.782                    & 0.831                     & 0.888                   & 0.903                    & 0.918                    & 0.835                    & 0.872                    & 0.86                     & 0.783                    & 0.866                     \\
\multicolumn{1}{|c|}{\textbf{Specificity}} & 0.999                    & 0.999                    & 0.999                    & 0.998                   & 0.999                    & 0.999                    & 0.998                    & 0.999                     & 0.999                   & 0.999                    & 0.999                    & 0.999                    & 0.998                    & 0.999                    & 0.999                    & 0.999                     \\
\multicolumn{1}{|c|}{\textbf{Precision}}   & 0.842                    & 0.88                     & 0.936                    & 0.909                   & 0.846                    & 0.873                    & 0.791                    & 0.868                     & 0.912                   & 0.931                    & 0.944                    & 0.914                    & 0.911                    & 0.983                    & 0.912                    & 0.929                     \\
\multicolumn{1}{|c|}{\textbf{DI}}          & 0.825                    & 0.887                    & 0.932                    & 0.874                   & 0.884                    & 0.772                    & 0.784                    & 0.851                     & 0.898                   & 0.925                    & 0.93                     & 0.877                    & 0.888                    & 0.909                    & 0.851                    & 0.897                     \\
\multicolumn{1}{|c|}{\textbf{S2S(mm)}}     & \multicolumn{1}{l}{1.152} & \multicolumn{1}{l}{1.130} & \multicolumn{1}{l}{1.084} & \multicolumn{1}{l}{1.401} & \multicolumn{1}{l}{1.825} & \multicolumn{1}{l}{1.977} & \multicolumn{1}{l}{2.287} & \multicolumn{1}{l|}{1.551} & \multicolumn{1}{l}{0.903} & \multicolumn{1}{l}{1.386} & \multicolumn{1}{l}{1.142} & \multicolumn{1}{l}{2.019} & \multicolumn{1}{l}{1.895} & \multicolumn{1}{l}{1.023} & \multicolumn{1}{l}{1.781} & \multicolumn{1}{l|}{1.450} \\ \hline
\end{tabular}
\end{adjustbox}
\vspace{-0.2cm}
\end{table}

\begin{figure}[!h]
\vspace{-0.4cm}
\centering
\includegraphics[height=5cm,width=0.85\textwidth]{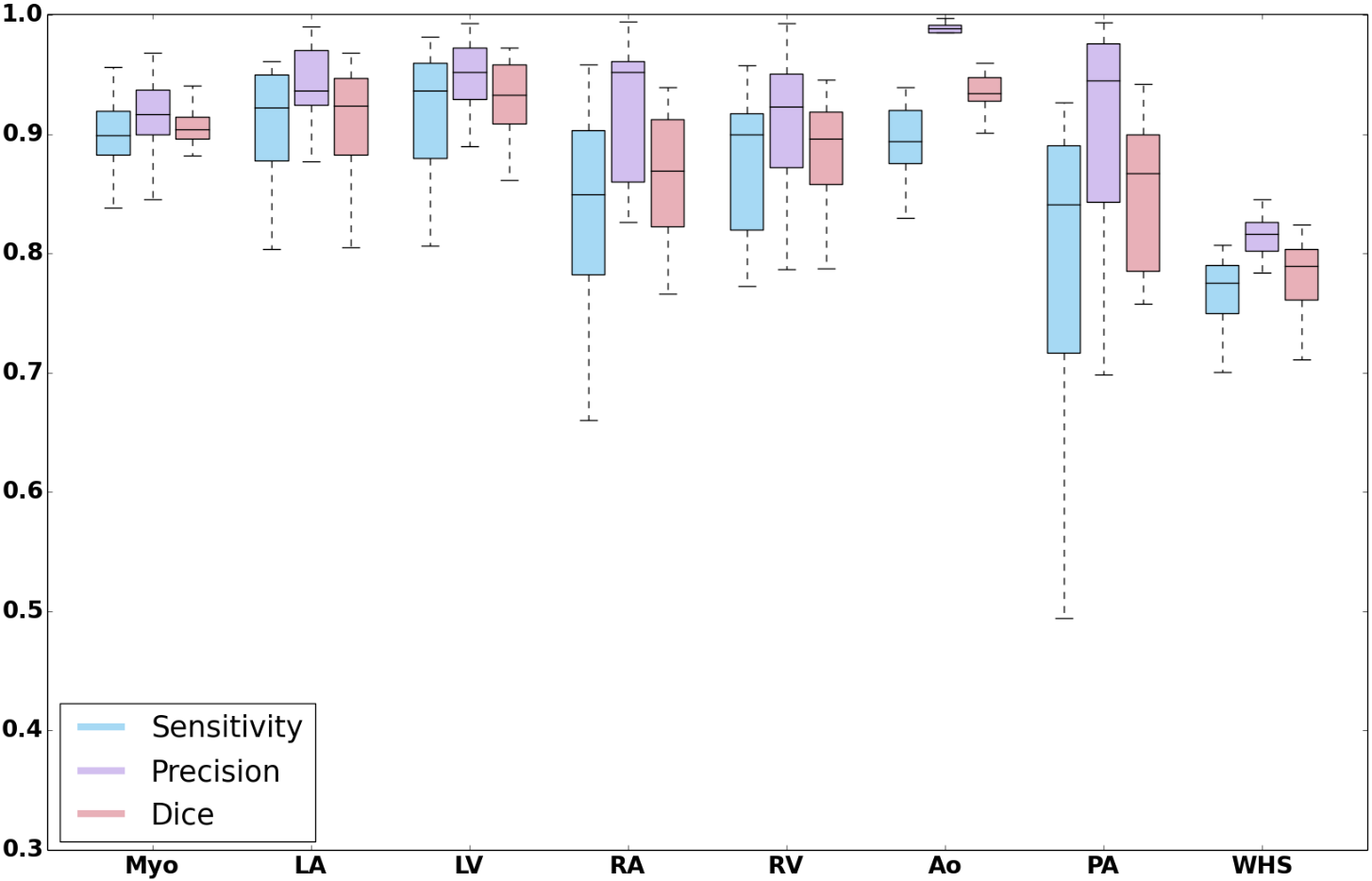}
\includegraphics[height=5cm,width=0.85\textwidth]{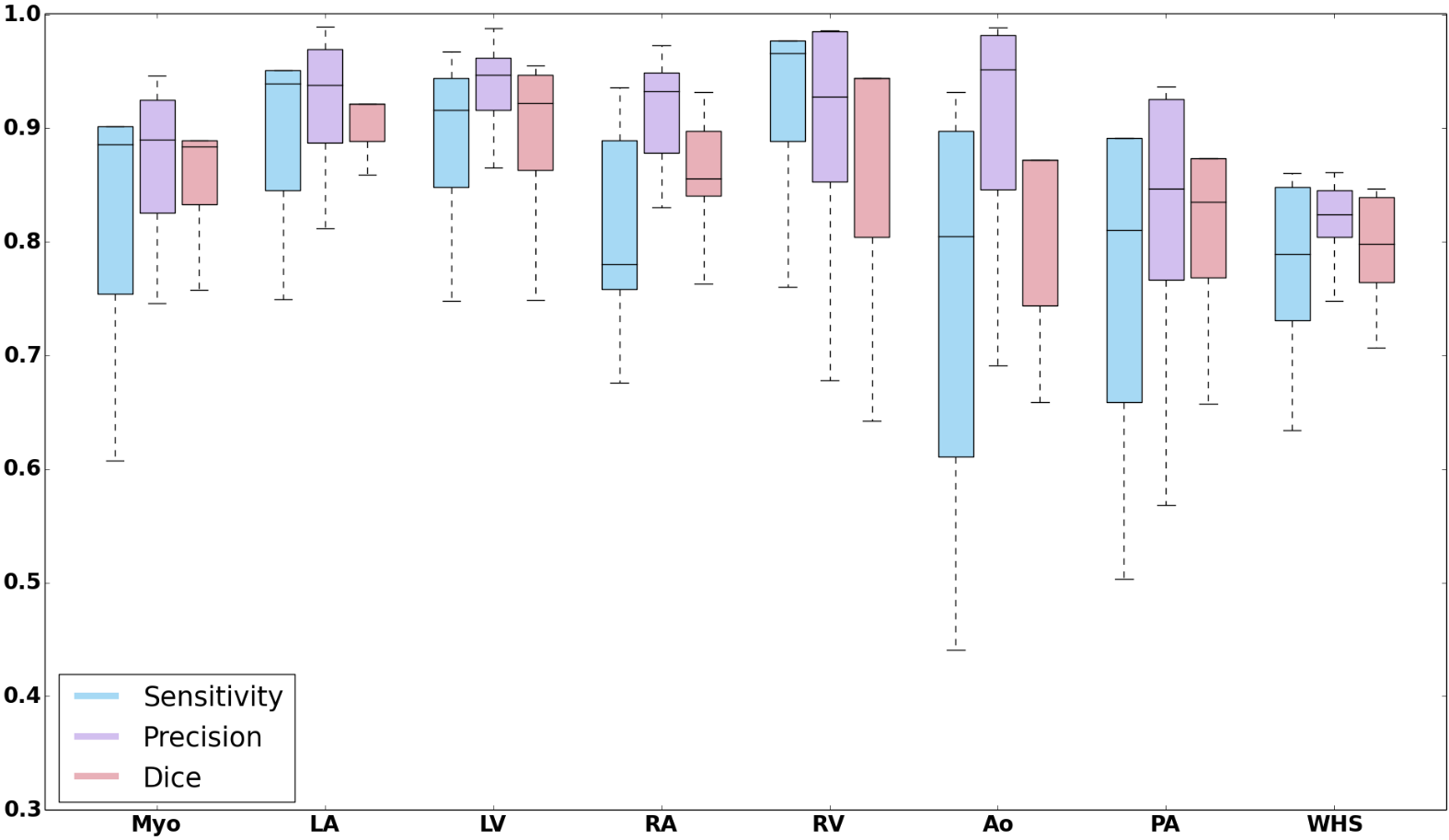}
\caption{Box plots for sensitivity, precision, and Dice index for each structure and WHS. Top figure is for CT dataset and bottom figure is for MR dataset \label{fig:boxplot}}
\vspace{-0.25cm}
\end{figure}

\noindent\textbf{Evaluation:} Five metrics were assessed: sensitivity, specificity, precision, dice index (DI), and surface to surface (S2S) distance. A summary of the findings for each structure and also for the whole heart are reported in Table 1. The WHS is the average of all structures. The box-plot for sensitivity, precision, and DI for both CT and MRI and for all structures are shown in Fig. \ref{fig:boxplot}. The qualitative results (including difficult cases for segmentation) for CT and MR modalities are illustrated in Fig. \ref{fig:qual}. Algorithms were implemented on the Nvidia TitanXp GPUs using Tensorflow~\cite{tensorflow}. The average time for segmenting the whole heart from the CT volume using three TitanXp GPUs was about 50 seconds. Segmenting using MR volume took about 17 seconds. For comparison, the time on the Intel Xeon Processor E5-2620 with 8 cores for CT images was about 30 minutes and for MR images was about 8 minutes. 

\section{Discussion and Conclusion}
\vspace{-0.2cm}
The main goal of the current study is to develop a framework for accurately segmenting the all cardiac substructures from both CT and MR images with high efficiency. The main strength of the proposed method is to train multiple CNNs from scratch and to allow an adaptive fusion strategy for information maximization in pixel labeling despite the limited data and hardware support. Our findings indicate that MO-MP-CNN can be used as an efficient tool to delineate cardiac structures with high precision, accuracy, and efficiency. 


Technically, one may question why we did not employ a completely 3D CNN approach instead of utilizing a multi-planar fusion of multiple 2D CNNs. As discussed in~\cite{mortazi2017}, the lack of a large number of 3D images restricts the  depth of  CNN training, which may highly likely result in sub-optimal implementation. Hence, training large number of 2D slices is much more feasible than utilizing 3D approach with the current setting. In the instance of plentiful GPU processing power and 3D imaging data, training would be optimized using a 3D CNN. 
 
Another limitation of our work stems from the use of the \textit{softmax} function in the last layer of the proposed network.  To explore whether the information loss due to class normalization in this step is significant, further research should be undertaken using information from the layer before the \textit{softmax} in fusion part and compared with the current system. Finally, further work is needed to establish comparative evaluation of different deep neural network approaches such as \textit{ResNet}, \textit{U-net}, and others.  While deeper networks are desirable to achieve  higher precision in segmentation tasks, lack of 3D data is a significant limitation for training such a system. Data augmentation and transfer learning have been shown to adequately address such challenges to a certain degree, but there is currently no research proving the optimality of such networks relative to the availability of data at hand. 

\vspace{-0.25cm}
\bibliographystyle{IEEEbib}
\bibliography{strings,refs}

\end{document}